\documentclass[10pt]{article} % For LaTeX2e
\usepackage[preprint]{tmlr}
\usepackage{amsmath,amsfonts,bm}

\def\eqref#1{equation~\ref{#1}}
\def\1{\bm{1}}

\DeclareMathAlphabet{\mathsfit}{\encodingdefault}{\sfdefault}{m}{sl}
\SetMathAlphabet{\mathsfit}{bold}{\encodingdefault}{\sfdefault}{bx}{n}

\usepackage{hyperref}
\usepackage{url}

\usepackage[ruled, linesnumbered]{algorithm2e}
\usepackage{amssymb}
\usepackage{booktabs}
\usepackage[pdftex]{graphicx}
\usepackage{stmaryrd}

\title{Meta Operator for Complex Query Answering on Knowledge Graphs}

\author{\name Hang Yin \email h-yin20@mails.tsinghua.edu.cn \\
       \addr Department of Mathematical Sciences\\
       Tsinghua University\\
       Beijing, China
       \AND
       \name Zihao Wang \email zwanggc@cse.ust.hk \\
       \addr Department of CSE\\
       HKUST\\
       Hong Kong SAR, China
       \AND
       \name Yangqiu Song \email yqsong@cse.ust.hk \\
       \addr Department of CSE\\
       HKUST\\
       Hong Kong SAR, China
       }

\newcommand{\entity}{\mathcal{E}}

\def\month{MM}  % Insert correct month for camera-ready version
\def\year{YYYY} % Insert correct year for camera-ready version
\def\openreview{\url{https://openreview.net/forum?id=XXXX}} % Insert correct link to OpenReview for camera-ready version

\begin{document}

\maketitle

\begin{abstract}
Knowledge graphs contain informative factual knowledge but are considered incomplete. To answer complex queries under incomplete knowledge, learning-based Complex Query Answering (CQA) models are proposed to directly learn from the query-answer samples to avoid the direct traversal of incomplete graph data. Existing works formulate the training of complex query answering models as multi-task learning and require a large number of training samples. In this work, we explore the compositional structure of complex queries and argue that the different logical operator types, rather than the different complex query types, are the key to improving generalizability. Accordingly, we propose a meta-learning algorithm to learn the meta-operators with limited data and adapt them to different instances of operators under various complex queries. Empirical results show that learning meta-operators is more effective than learning original CQA or meta-CQA models.
\end{abstract}

\section{Introduction}
Knowledge graphs (KG) encode factual knowledge of the world as a collection of relational edges between entities and may serve as a knowledge base for many downstream AI-related tasks~\citep{ren_fact_2023}. 
However, real-world KGs like NELL~\citep{carlson_toward_2010} are auto-generated, making them both massive and incomplete. Incomplete knowledge brings fundamental challenges to the tasks that can not be solved by traditional logical reasoning~\citep{arakelyan_complex_2020}, which is also known as Open World Assumption (OWA)~\citep{libkin_open_2009}.

\begin{figure}[t]
    \centering
    \includegraphics[width=1\linewidth]{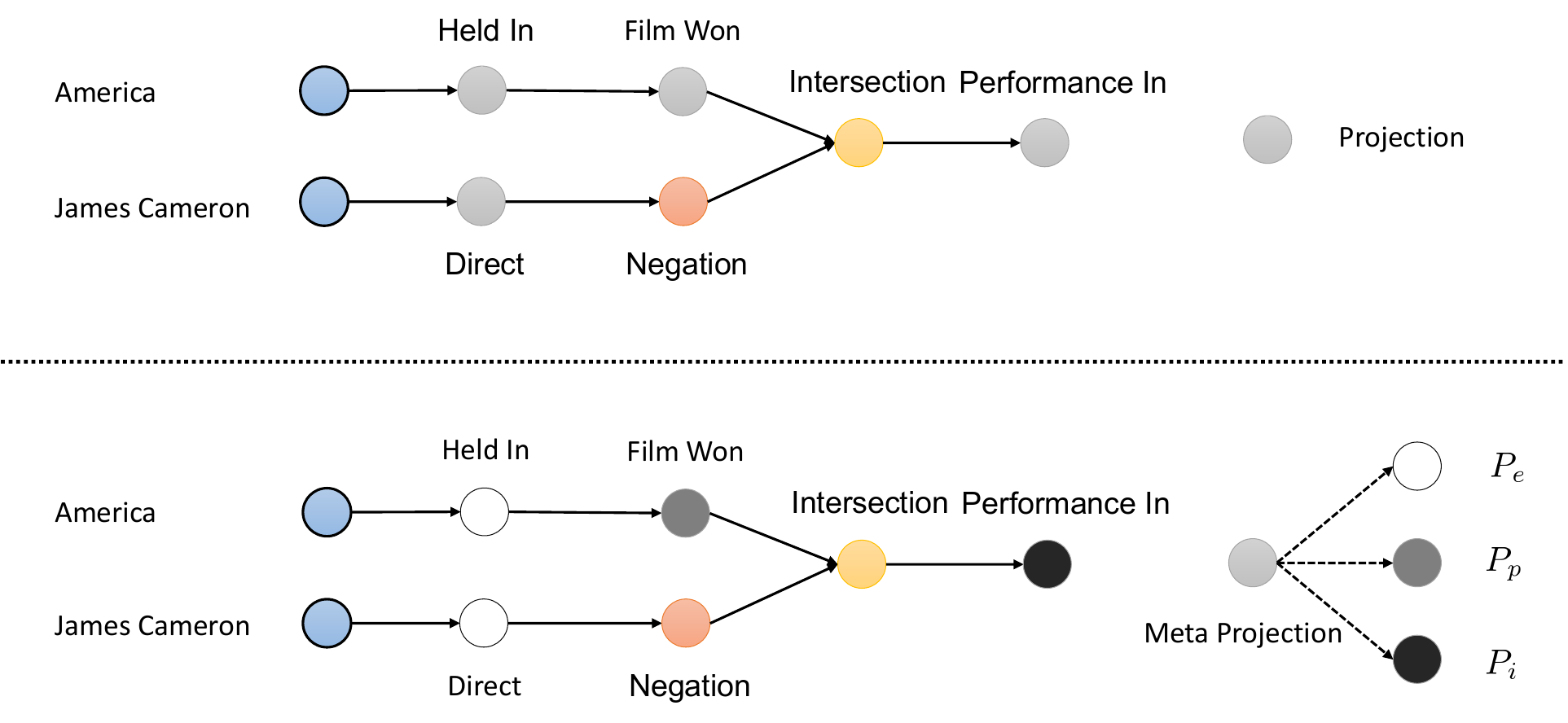}
    \caption{Top: Original computation tree of the query ``Who has performed in such a movie, directed by James Cameron but has not won any award held in America?''. Bottom: MAMO computation tree, where we show that the meta projection is adapted to three different operator types according to the ``input'' categorization.}~\label{fig:MAMO computation tree}
\end{figure}

An essential task on KG is complex query answering (CQA)~\citep{wang_benchmarking_2021,yin_rethinking_2023}, which aims to answer complex queries using the partially observed KG and has been proven to be very useful in knowledge base question answering~\citep{ren_lego_2021}, fact ranking~\citep{ren_fact_2023}o.
Such complex queries can be formally represented by first-order logic (FOL) formulas, involving three types of logical operators: conjunction($\land$), disjunction($\lor$), and negation($\lnot$). 
``Who has performed in such a movie, directed by James Cameron but has not won any award held in America?'' is an example of the complex query, as shown in Figure~\ref{fig:MAMO computation tree}. 
However, due to the OWA, direct logical operator execution can not derive the whole correct answer.

Therefore, query embedding methods are proposed ~\citep{arakelyan_complex_2020,choudhary_probabilistic_2021,zhang_cone_2021,amayuelas_neural_2021,bai_complex_2023,bai_knowledge_2023,wang_wasserstein-fisher-rao_2023,fei_soft_2024} to leverage machine learning models to infer the missing knowledge.
These methods embed entities and queries in low-dimensional space and treat logical operator execution as computations on these embeddings.
Compared to traditional methods, query embedding methods train neural models on queries over the observed knowledge graphs and allow fast inference compared to the symbolic search methods~\citep{ren_query2box_2020,yin_textefo_k-cqa_2023}. Most importantly, query embedding methods are possibly equipped with out-of-distribution generalizability since the unseen and arbitrarily complex query types are also composed of seen atomic logic operators. Empirical results~\citep{ren_beta_2020} have shown that CQA models trained on limited query types are capable of answering complex queries of unseen types.

% Although CQA models have shown their capability of combinatorial generalization to some extent, that does not mean that CQA models can achieve an acceptable performance even if they have only been trained on the simplest formulas. In contrast, to alleviate the problem of combinatorial generalizability, the previous solution was to add various formulas to the training dataset, nowadays standard dataset provided by \citet{ren2020beta} includes several hundred thousand queries for 10 different formulas to help the model achieve satisfactory performance. \citet{chen2022fuzzy} has conducted an experiment with one-hop queries being the only available training data and found out that existing CQA models surfers from severe drawbacks, even though theoretically, the one-hop queries already carry all information needed. This solution is not satisfactory, since computing embedding for complex formulas gets more and more computation intensive as the computation graph goes deeper and more complicated. There are also situations where training data for complex queries are rare or hard to access. 

Although CQA models are believed to be combinatorially generalizable, further empirical evaluation~\citep{chen_fuzzy_2022} shows that with the one-hop query being the only available training data, existing CQA models suffer from severe drawbacks. This finding indicates the fact that a large number of training samples of complex multi-hop queries, such as the training dataset proposed by~\citet{ren_beta_2020}, is essential for the performance of CQA models.  
Liu~\citep{liu_neural-answering_2021} shows that the errors by one-hop queries can be accumulated when answering multi-hop queries. These observations reveal that simple one-hop queries are insufficient to train those query embedding methods even though one-hop queries have already contained all knowledge that can be retrieved from the observed knowledge graph in theory.
Meanwhile, Wang~\citep{wang_benchmarking_2021} showed that simply adding more multi-hop queries to the training dataset may not be the best solution as it can harm the performance of one-hop queries.
In the terminology of multi-task learning, it is straightforward to regard the query types as different tasks. Then, the phenomenon discussed above shows that the interaction mechanisms between different tasks are complicated and not well understood. Therefore, current learning algorithms might produce sub-optimal models.

% What's more, the data offered by different formulas may 'conflict' with each other. This phenomenon is firstly observed by \citet{liu2020k}, who addressed this issue by a term, 'cascading error' for models like query2box~\citep{ren2020query2box} and TransE~\citep{bordes2013translating}, as the latter two treat projection as a simple linear transformation, making these models fail to generalize to longer, more complicated formulas even though they perform well in one-hop queries. A recent and more detailed experiment made by \citet{wang2021benchmarking} offers a novel observation that more training query types do not necessarily lead to better performance. Simply adding multi-hop queries to the training set improve their own test performance at the cost of impairing the performance of one-hop queries. This all suggests that simply adding more formulas to the training data without considering their interactions, though being a standard treatment by now, may not be an optimal way to train CQA models.  
In this paper, we propose to coordinate the training of query embedding methods by meta-learning, especially under the few-shot learning scenario, specifically, when the observed knowledge graphs remain unchanged but the size of the training query/ query type is restricted. Instead of directly applying meta-learning to adapt query embedding models into different complex query types, we propose to explore the operator structure inside the complex query types. In this work, we propose a Model Agnostic Meta Operator (MAMO) algorithm to learn meta logical operators and then adapt to the specific operator instance inside arbitrary complex query types. We also explore different ways to categorize the operator types in a complex query. Our experiments show the effectiveness of our model in the few-shot setting, and our results imply that the logical operator types, rather than the composed logical query types are the key to combinatorial generalizability.

% In this paper, we propose to coordinate the training of query embedding methods by meta-learning under the few-shot learning scenario. Instead of directly applying meta-learning to adapt query embedding models into different complex query types, we propose to explore the operator structure inside the complex query. In this work, we propose a Model Agnostic Meta Operator algorithm to learn meta logical operators and then adapt to the specific operator instance inside arbitrary complex queries. Our experiments show the effectiveness of our model in the few shot setting, and our results implies that the logical operators, rather than the composed logical queries is the key of meta learning

% To solve this challenge, we aim to get a decent result on a new formula as long as all of its operators have been trained even though the number of its own query can be strictly limited. 
% In this paper, we propose a novel algorithm called Model Agnostic Meta Operator(MAMO), which is inspired by one of the most influential and widely adopted meta-learning algorithms, Model Agnostic Meta-Learning(MAML)~\citep{finn2017MAML}. To the best of our knowledge, we are the first to leverage meta-learning method for complex query answering on KG.

\section{Related Work}
\subsection{Query Embedding-Based CQA Models over KGs} 

These methods embed entities and queries as geometric objects~\citep{zhang_cone_2021,bai_query2particles_2022} or probabilistic distribution~\citep{ren_beta_2020,choudhary_probabilistic_2021} or even fuzzy logic~\citep{chen_fuzzy_2022,yin_existential_2023} in the low dimensional space, where embeddings of query and its answer entities are required to be close to each other. Although they use various parametrizations for logical operators and query embeddings, they all share the same dataset, and their training procedures are nearly the same~\citep{wang_logical_2022}. This kind of consistency, on the other hand, means that training models in a scenario where the number of data is strictly limited or different ways to train query embedding models have not been studied in detail yet.

\subsection{Meta-learning for Few Shot Setting}

MAML~\citep{finn_model-agnostic_2017} is a highly popular meta-learning algorithm that is designed for few-shot learning, it has shown great success on many benchmark~\citep{vinyals_matching_2016} of few-shot learning problems. In addition, it has inspired lots of variants~\citep{raghu_rapid_2019} in the big family of optimization-based meta-learning algorithms, exploring fields like few shot learning on graphs~\citep{zhou_meta-gnn_2019}. However, to the best of our knowledge, it has not been applied to the domain of complex query answering yet.

\section{Preliminaries}

\subsection{Knowledge Graphs}
Given a set $\entity$ of entities (nodes) and a set $\mathcal{R}$ of relations (edge types), a knowledge graph  $\mathcal{KG} = \{(r,h,t)\}$, where each element is a factual triplet $(v_h, r, v_t)\in \entity \times \mathcal{R} \times \entity$, $v_h,v_t$ are known as head entity, tail entity, respectively, and the $r$ represents the relation between them.
\subsection{EFO-1 Query}
The task we focus on is complex query answering, which can be expressed as a subset of Existential First Order Logic Query with a single free variable (EFO-1 query). We assume that $\{x_i\}_{i=1}^{n}$ is the non variable anchor entity set, $y_1,\dots ,y_k$ denote existentially bounded variables, and $y_?$ is the single target variable. The Disjunctive Normal Form (DNF) of the EFO-1 query can be formulated as follows:
$$Q(y_?)=\exists y_1 \exists y_2 \dots \exists y_k (v_{11}\land \dots \land v_{1N_1})\lor \dots \lor(v_{M}\land \dots \land v_{MN_M})$$
where each $v_{ij}$ is one of the following: $r(x_a, y_b)$, $\lnot r(x_a, y_b)$, $r(y_a, y_b)$, or $\lnot r(y_a, y_b)$,  in which $r\in \mathcal{R}$, $x_a \in \{x_i\}$, $y_a, y_b \in\{y_?,y_1,\dots,y_k\}$, $y_a \neq y_b$. The predicate $r$ is induced by relation on the knowledge graph naturally: 
$$r(v_h, v_t)=\text{True}  \iff  (v_h, r, v_t)\in \mathcal{KG}$$

Our goal is to answer the logic queries $Q$, namely, to find the answer set $\llbracket Q \rrbracket\subset \entity$ for the free variable $y_?$, such that $v\in \llbracket Q \rrbracket \iff Q[v] = \text{True}$.

Additionally, if we make a further restriction that the EFO-1 query does not contain any negation within it, we call this subset Existential Positive First Order Logic Query (EPFO query). 
\subsection{Computation Tree}
To estimate the embedding of the target answer, existing methods~\citep{wang_benchmarking_2021} utilize the idea of representing the query as a computation tree,  where nodes represent entity sets, and edges represent logical operations over entity sets. Specifically, the leaf nodes are anchor nodes and the root of the tree naturally represents the target answer, the mapping along each edge applies a certain logic operator, including:

\begin{itemize}
    \item Set Projection: Given set of entities $A$, relation $r$, compute ${\{b\in\entity |~ \exists a\in A,~r(a,b)\in \mathcal{KG}\}}$ 
    \item Set Intersection: Given two sets of entities,$A, B$ return $A\cap B$.
    \item Set Union: Given two sets of entities, $A, B$ return $A\cup B$.
    \item Set Complement(Negation): Given set of entities $A$, return $\entity-A$.
\end{itemize}
    
More details can also be found in~\citet{ren_fact_2023}. The so-called query embedding method therefore parametrizes these logical operations and allows for computing the embedding of the answer by recursively computing embeddings from the anchor nodes in a bottom-up manner~\citep{ren_beta_2020}. We offer an example in Figure~\ref{fig:MAMO computation tree}. The example contains operators including projection, intersection (related to logical conjunction), and negation. For union(logical disjunction), it can be answered by aggregating all its conjunctive sub-queries~\citep{ren_query2box_2020,wang_logical_2023}.

\subsection{Meta Learning Terminology} \label{sec:meta learning preliminary}

Here we introduce the basic settings and corresponding terminologies of meta-learning. Meta-training differs from conventional supervised machine learning in the way that it aims to learn the parameter that performs well in many different tasks~\citep{hospedales_meta-learning_2020}. To fulfill this goal, each training step of meta-learning is split into two stages: a \textbf{inner} stage, and a \textbf{outer} stage~\citep{zhou_meta-gnn_2019,hospedales_meta-learning_2020}. Additionally, the training data is also split into two parts: the \textbf{support data} $S$ in the inner stage, and the \textbf{target data} $T$ in the outer stage. We use $L_S$ to represent the loss function in the data $S$. In the inner stage, the goal of meta-learning is to do the \textbf{adaptation}: it adapts the meta parameters to any specific task in the support data. In the outer stage, we use the adapted parameters to compute \textbf{outer loss} $L_{outer}$ of the target data and then update the original meta-parameters accordingly, this stage is also known as meta-optimization in MAML. This training method explicitly learns the meta parameters that can adapt to any specific task easily. 

%\subsection{Formula and query}
%In this paper, we follow the terminology used in \citet{wang2021benchmarking}, calling query %type as formula, and grounded query just as query. Additionally, we also use the lisp %language it proposed as it is more systematic and comprehensive.

\section{Methodology}
% In light of the observation that the interaction within data of different query types is complex and unknown as we mentioned before,
It has been found by previous empirical observation that, the interaction within data of different query types is complex and unknown, making it hard to learn the operator that performs well in all query types~\citep{wang_benchmarking_2021}.

To remedy this problem, we proposed the Model Agnostic Meta Operator (MAMO) to avoid the direct learning of the specific operator. Instead, MAMO learns a meta operator that can achieve satisfactory performance at any specific instance of this operator. The critical point of MAMO is that the different appearances of the same operator rather than different query types play a pivotal role in promoting the combinatorial generalizability of CQA models. Therefore, we argue that our MAMO algorithm is the proper way to apply meta-learning to CQA models, rather than the alternative, namely applying MAML at the query type level, which is explained in Section~\ref{sec:MAML explained}.

\begin{table}[t] 
\caption{Details of different categorizations of projections.} \label{tab: positional operator}
\centering
\scriptsize
\begin{tabular}{lll}
\toprule
Name             & How to Categorization                                            & All Operator Types\\ \midrule
Root (R) & Distance to the root node.           & $\{1,2,\dots,\infty \}$ \\
Leaf (L) & Distance to the nearest leaf node. & $\{1,2,\dots,\infty\}$ \\ \midrule
Input (I) & Input node type.                                           & Projection, Intersection, Entity                                  \\
Output (O) & Output node type. & Projection, Intersection, Negation, Answer                                                                \\  \midrule
Binary Input(BI) & Whether the input node is entity. & Entity, Non-entity. \\
Binary Output(BO) & Whether the output node is answer. & Answer, Non-answer. \\
\bottomrule
\end{tabular}
\vskip-2em
\end{table}

\subsection{Operator Types}

First of all, we need to decide whether an operator should be learned by the meta-learning method or not. Previous empirical observation has suggested that projection is the most critical learnable operator for model performance~\citep{chen_fuzzy_2022}. In the following paper, for simplicity, we constrain our discussion so that projection is the only operator that needs to be learned by meta-learning. We note that applying the MAMO algorithm for other operators such as intersection is easy and straight-forward.

To distinguish the instance of projection, motivated by~\citet{bogin_unobserved_2022}, we take the position of projection within the computation tree into account, thus resulting in six different kinds of categorizations for operator types, which are explained in detail in Table~\ref{tab: positional operator}.

In the following discussion, for simplicity, we will illustrate our algorithm by using the Input (I) categorization as an example to categorize different projection operators. This method categorizes projection by the input of the projection, since the input for projection can only be projection, intersection, and entity, we notate the three subcategories of projection as $P_p, P_i, P_e$, where the subscript corresponds to the operator type. Figure~\ref{fig:MAMO computation tree} shows an example of classifying operator type of projection in one computation tree. 

For models that share parameters for all appearances of projection, we denote the shared parameters of projection as $\theta$ and the other learnable parameters of models as $\phi$. However, when the operator type for projection is introduced, we need different parameters for different operator types, we reuse $\theta$ to represent the parameter for meta projection and use $\theta_p, \theta_i, \theta_e$ to represent the parameters for $P_p, P_i, P_e$ correspondingly. Then on a batch of data $D$, the original loss function $L_D(\theta, \phi)$ should be extended to $L_D(\theta_p, \theta_i, \theta_e, \phi)$, we separate these three parameters in the loss function to facilitate the adaptation of each one.

\subsection{Learning and Adapting Meta Operators}

Following our introduction in Section~\ref{sec:meta learning preliminary}, we describe our MAMO algorithm in the two stages: in the meta-training stage, our goal is to adapt the parameters $\theta$ of meta projection into the parameters of the three different operator types: $\theta_p, \theta_i, \theta_e$; in the meta-testing stage, we use the three adapted parameters to update the initial parameters $\theta$ of the meta projection. This training method requires us to split the data into support set $S$ and target set $T$, used for meta-training and meta-testing, respectively.

In meta-training stage, we use gradient descent in the support set $S$ to adapt the meta parameter $\theta$ to the type-specific $\theta_p, \theta_i, \theta_e$. To get an accurate gradient for a specific operator type, for example, $P_p$, we only need to consider those queries that contain an instance of $P_p$, thus getting $S_{P_p}\subset S$. Given the adaptation learning rate $\alpha$, the adaptation should be 
\begin{equation}\label{eq: inner adapt}
	\theta_{p}=\theta - \alpha *\nabla_{\theta_p} L_{S_{p_p}}\bigg(\theta_{p}, \theta_e=\theta, \theta_i=\theta, \phi\bigg)\bigg|_{\theta_p=\theta}
\end{equation}

In the meta-testing stage, after computing all $\theta_p, \theta_i, \theta_e$ in the meta-training, we can aggregate them and compute the outer loss in the target set $T$
\begin{equation}
    L_{outer}=L_T(\theta_{p}, \theta_{i}, \theta_{e}, \phi)
\end{equation}

Then, the outer loss is fed into the Adam optimizer to update $\theta, \phi$ accordingly. In this way, we learn the meta-parameter $\theta$ that can adapt to any specific operator type easily. In testing, we just do the adaptation and use the adapted parameters $\theta_{p}, \theta_{i}, \theta_{e}$ for inference.

We also show the pseudo-code for the MAMO in the Algorithm~\ref{alg: algorithm}.

\begin{algorithm}
	\caption{MAMO for CQA models}
	\label{alg: algorithm}
	\KwIn{(1)Dataset $D$, (2)adaptation learning rate $\alpha$, (3)the set of all operators that need to be learned by MAML $M$, (4)the categorization method.} 
        \KwOut{Trained parameters for CQA models}
	Initialize all model parameters, for an operator $O \in M$, we denote its parameters as $\theta_O$, and we denote all other parameters as $\phi$\;
    Compute all operator types for every operator $O\in M$, denoted the set of operator types of $O$ as $M_O$\;
	\While{not done} {
		Sample batch of queries $S\in D$ as support set\;
            \For{operator $O\in M$}{
		      \For{operator type $O_t\in M_O$} {
				Select those queries that contain at least one instance of $O_t$ in S, getting set of queries $S_{O_t} \subset S$\;
				%Freezing the appearance of all other positional types of O\;
				Compute the adapted parameters for $\theta_{O_t}$ on queries $S_{O_t}$ by equation \ref{eq: inner adapt} \;
				% Evaluate inner loss for all formulas in $q_n$ that contain operator type t: $L_t(\theta)$ as in \ref{eq: inner adapt}\;
				% Compute adapted parameters for specific operator type $\theta_{t}=\theta - \alpha *\nabla_{\theta} L_t(\theta)$\;
		      }
            }
		Sample batch of queries $T\in D$ as the target set\;
		Compute outer loss by aggregating all $\{\theta_{O_t}\}_{O\in M,O_t\in M_O}$: $L_{outer}=L_T(\{\theta_{O_t}\}_{O\in M,O_t\in M_O}, \phi)$\;
		Update original $\theta$ and $\phi$ using the $L_{outer}$;
	}
\vskip-.5em
\end{algorithm}

\subsection{Math Analysis of MAMO}\label{app:math}
In this section, we make a math analysis of MAMO to show its interpretability.
Consider we apply MAMO to parameter $\theta$, the loss function in a given query $Q$ where $\theta$ has appeared for $n$ times is defined to be:
$$L=L_Q(\theta_1, \dots, \theta_n, \phi)$$ where each $\theta_i$ denotes the specific appearance of the operator in the computation tree of the query $Q$, and $\phi$ is other parameters for the CQA models.
The original model computes gradient to update the $\theta$ as: $$\nabla_{\theta}L= \Sigma_{i=1}^{n} \frac{\partial}{\partial \theta_i}L_Q\left(\theta_1\bigg|_{\theta_1=\theta}, \dots,\theta_i\bigg|_{\theta_i=\theta}, \dots, \theta_n\bigg|_{\theta_n=\theta}, \phi\right)$$
For any categorization, it splits the appearance of this operator into k different operator types, thus creating k subsets $T_1, \dots, T_k$ as a partition of $\{1, \dots, n\}$: $$\cup_{i=1}^k T_i=\{1, \dots, n\}, i\neq j \implies T_i \cap T_j=\varnothing $$
Then in MAMO, the gradient for the operator type corresponding to $T_i$ as:
$$\Sigma_{j \in T_i}\nabla_{\theta_{j}}L = \Sigma_{j\in T_i}\frac{\partial}{\partial \theta_j}L_Q\left(\theta_1\bigg|_{\theta_1=\theta}, \dots,\theta_j\bigg|_{\theta_j=\theta}, \dots, \theta_n\bigg|_{\theta_n=\theta}, \phi\right)$$

And we immediately find out that:
$$\Sigma_{i=1}^{k}\Sigma_{j \in T_i}\nabla_{\theta_{j}}L = \Sigma_{i=1}^{n}\nabla_{\theta_i}L$$

By this finding, we show that the adaptation from meta operator to specific operator types is just splitting the gradient in the initial training. The categorization decides how to regroup the gradient. That explains the effectiveness of our MAMO algorithm since it computes the gradient in a more detailed way.

\section{Experiments} \label{sec:experiment}
In this section, we evaluate our new proposed algorithm applied to existing CQA models. Our experiments are mainly done by using LogicE~\citep{luus_logic_2021} as the backbone model, to show that our algorithm is model-agnostic, we also show the outcome of choosing FuzzQE~\citep{chen_fuzzy_2022} and ConE~\citep{zhang_cone_2021} as the backbone CQA model. We use the standard metric, Mean Reciprocal Rank(MRR) as the indicator of model performance, the higher the better. Our code and data will be released after acceptance.

\begin{table}[t]
\centering
\caption{The MRR result(\%) of LogicE in the multi-hop queries.} \label{tab:onlyP LogicE}
\begin{tabular}{llllllll}
\toprule
Algorithm & 1p    & 2p   & 3p   & 4p   & 5p   & 6p   & AVG.  \\
\midrule
LogicE  & 40.25 & 5.75 & 4.66 & 4.21 & 3.87 & 3.81 & 10.42 \\
\midrule
MAML      & 39.87 & 6.21 & 4.41 & 3.87 & 3.84 & 2.88 & 10.18 \\
\midrule
MAMO(R)   & 40.70 & 6.50 & 4.86 & 4.30 & 4.24 & 4.20 & 10.80 \\
MAMO(L)   & 40.64 & 6.47 & 4.93 & 4.38 & 4.26 & 4.18 & 10.81 \\
MAMO(I) & 40.66          & 6.50          & 4.90          & \textbf{4.39} & \textbf{4.43} & 4.22          & 10.85          \\
MAMO(O) & \textbf{40.83} & \textbf{6.60} & \textbf{4.95} & 4.35 & 4.40          & \textbf{4.23} & \textbf{10.89}
\\
\bottomrule
\end{tabular}
%\vskip-2em
\end{table}
\subsection{Experiment Setup}
\subsubsection{Baselines}\label{sec:MAML explained}
We compare our MAMO algorithm with two baselines: (1) The original backbone CQA model. (2) MAML directly applied to CQA models. Here we briefly describe the training procedure of the directly applied MAML as follows.

We straightforwardly treat different query types as different tasks: in every training step, we need to sample a task, namely a query type $f$, and sample two batches of data within this query type $f$ as support set $S$ and target set $T$.

In the meta-training stage, given the loss function $L$, the meta-parameters for the whole model, denoted as $\theta$, should be adapted to $\theta_f$, corresponding to the query type $f$:
$$\theta_f = \theta - \nabla_{\theta}L_S(\theta)$$

In the meta-testing stage, the outer loss is computed in the target data to update the original $\theta$:
$$L_{outer}=L_T(\theta_f)=L_T(\theta - \nabla_{\theta}L_S(\theta))$$
$$\theta \leftarrow \text{UPDATE}(\theta, L_{outer})$$
\subsubsection{Dataset}
Our experiments are conducted in a few-shot setting. Specifically, we derive our dataset from the standard dataset proposed by~\citet{ren_beta_2020} on the knowledge graph FB15k-237~\citep{toutanova_observed_2015} since most CQA models have adopted this dataset as their training dataset, and FB15k-237 is reported to be the most challenging knowledge graph. This dataset contains 14 different query types: 
1p/2p/3p/2i/3i/ip/pi/2u/up/2in/3in/inp/pin/pni. We note that ip/pi/2u/up are only used for evaluation, which tests the combinatorial generalizability of the model. We retain only 0.1\% of the training samples except for the one-hop queries since one-hop queries are directly related to the observed knowledge. Based on this few-shot dataset, we provide three different experiment settings: (1) Multi-hop: The model is trained on the 1p/2p/3p queries and evaluated on the 1p/2p/3p/4p/5p/6p queries\footnote{1p is the one-hop query, 2p is the two-hop query, and so on. The data of 4p/5p/6p is not included in the previous BetaE dataset and is sampled by us.}. (2) EFO-1: We use all training query types of the BetaE dataset and evaluated all query types. (3) EPFO: The models are only trained and evaluated on the EPFO part of the BetaE dataset. 

The reason we consider the EPFO queries even though it is a subset of EFO-1 is that the negated information can be unobserved rather than definitely wrong. We give an example for illustration: when considering the query ``Find a movie that has won the Oscar award but not directed by James Cameron'', meanwhile in the training graph the link shows ``Titanic is directed by James Cameron'' is missing, this will lead to the consequence that Titanic will count as an answer to this query while it is actually not. These negation queries provide us with ``fake answers'' and thus we mainly focus on the EPFO queries to exclude these problematic queries.

\subsubsection{Hyperparameters Setting}

The parameters of the backbone model are kept for fairness, thus, we only discuss the new parameters introduced by meta-learning.

In the training period, the support set for MAMO and MAML is both 32, the adaptation learning rate is 0.016 for LogicE, 0.1 for FuzzQE, and 0.008 for ConE, same for MAMO and MAML, which are chosen from $\{0.004,0.008,0.016,0.03,0.05,0.1,0.2\}$. In inference time, the support set is  10 for both MAMO and MAML, the fine-tune learning rate is one-fourth of the corresponding adaptation learning rate for both algorithms, and they are all fine-tuned by five steps. 

However, we note that MAML has more data in the inference time since we feed it with out-of-distribution data.
\
%(1) We train and evaluate on only the 1p/2p/3p\footnote{1p stands for one-hop query, 2p is two-hop, and 3p is three-hop} queries since these query types are where the projection operation dominates and our categorization principles are designed for projection operators.

\begin{table}[t]
\centering
\caption{MRR result(\%) of LogicE in all EPFO queries.} \label{tab:EPFO formulas LogicE}
\begin{tabular}{lllllllllll}
\toprule
Algorithm & 1p    & 2p   & 3p   & 2i    & 3i    & ip   & pi    & 2u    & up   & AVG.  \\ \midrule
LogicE  & 40.73 & 5.74 & 4.54 & 27.74 & 37.66 & 8.21 & 17.59 & 12.61 & 4.46 & 17.70 \\ \midrule
MAML      & 40.32 & 6.44 & 4.81 & 25.38 & 34.22 & 7.84 & 16.37 & 11.94 & 5.48 & 16.98 \\ \midrule
MAMO(R)    & 40.85 & 6.17 & 4.51 & 29.66 & 40.11 & 8.30 & 18.24 & 13.12 & 4.76 & 18.41 \\
MAMO(L)    & 40.84 & 6.05 & 4.61 & 29.65 & 40.00 & 8.71 & 18.03 & 13.04 & 4.74 & 18.41 \\
MAMO(I)   & 40.97 & 6.21 & 4.77 & 29.46 & 39.86 & 8.70 & 18.09 & 13.15 & 4.90 & 18.46 \\ 
MAMO(O) &  \textbf{41.11} &  \textbf{6.50} &  \textbf{4.83} &  \textbf{30.04} &  39.82 &  \textbf{8.79} &  \textbf{18.58} & 13.21 &  \textbf{5.14} &  \textbf{18.67} \\
MAMO(BO) & 40.93 &  6.13 &  4.69 &  29.85 &  \textbf{40.32} &  8.57 &  18.27 &  \textbf{13.39} &  4.83 &  18.55 \\
\bottomrule
\end{tabular}
\end{table}

\begin{table}[t]
\centering
\scriptsize
\caption{The MRR result(\%) of LogicE in all EFO-1 queries.} \label{tab:all formulas LogicE}
\begin{tabular}{llllllllllllllll}
\toprule
Algorithm & 1p & 2p & 3p & 2i & 3i & ip & pi & 2in & 3in & inp & pin & pni & 2u & up & AVG. \\ \midrule
LogicE & 41.01 & 5.58 & 4.62 & 27.58 & 37.86 & 7.50 & 17.63 & 1.51 & 4.18 & 2.87 & 0.61 & 1.61 & 12.70 & 4.47 & 12.12 \\ \midrule
MAML & 39.90 & \textbf{6.30} & 4.46 & 24.76 & 33.51 & 7.40 & 15.35 & 0.95 & 2.48 & 3.33 & 0.60 & 0.94 & 11.42 & 5.00 & 11.17 \\ \midrule
MAMO(R) & 40.91 & 5.94 & 4.65 & 29.50 & 40.04 & 8.68 & 17.91 & 1.91 & 4.58 & 3.51 & 0.83 & 1.77 & \textbf{13.27} & 4.72 & 12.73 \\ 
MAMO(L) & \textbf{41.07} & 6.04 & \textbf{4.72} & 29.40 & 39.92 & \textbf{8.85} & 18.01 & 1.89 & 4.61 & 3.42 & 0.82 & 1.79 & 12.94 & 4.90 & 12.74 \\
MAMO(I) & \textbf{41.07} & 6.01 & 4.59 & 29.15 & 39.76 & 8.42 & 18.41 & 1.68 & 4.74 & \textbf{3.55} & 0.71 & 1.60 & 12.93 & 4.96 & 12.68 \\
MAMO(O) & 40.96 & 6.12 & 4.68 & 29.73 & \textbf{40.32} & 8.61 & 18.41 & \textbf{1.98} & \textbf{4.83} & 3.50 & 0.75 & \textbf{1.96} & 12.75 & 4.91 & 12.82 \\
MAMO(BI) & 41.02 & 6.12 & 4.41 & 29.57 & 40.14 & 8.53 & 18.19 & \textbf{1.98} & 4.73 & \textbf{3.55} & 0.76 & 1.71 & 13.16 & 4.85 & 12.76 \\
MAMO(BO) & 40.92 & 6.16 & 4.54 & \textbf{30.06} & 40.15 & 8.82 & \textbf{18.45} & 1.80 & 4.70 & 3.42 & \textbf{0.88} & 1.80 & 12.99 & \textbf{5.03} & \textbf{12.84} \\ \bottomrule
\end{tabular}
%\vskip-2em
\end{table}

\subsection{Results on Multi-Hop Queries}
We first look into the experiment of LogicE in the multi-hop settings to test whether the performance of projection can be improved by our MAMO algorithm. The result is shown in Table~\ref{tab:onlyP LogicE}, where the letter in the bracket denotes the categorization applied. The categorization of Binary Input and Binary Output is omitted since they are the same as Input and Output, respectively. 

The performance of MAML is unsteady and on average slightly worse than the performance of original LogicE, while the MAMO succeeds to outperform the original model in all four variants and all six query types, suggesting that applying meta-learning at the operator level is the better alternative to train projection. Among the results of MAMO, we compare the Input(I) with the Output(O), the Input performs better in longer queries 4p/5p/6p while impairing the performance in shorter queries such as 1p and 2p: this phenomenon also verifies our previous claim that the inner interaction within different query types is complex and unknown, even we only consider the multi-hop queries, we might find that a model that performs poorly in one-hop query might turn out to generalize well in multi-hop queries.

\begin{table}[t]
\centering
\caption{The MRR result(\%) of FuzzQE in all EPFO query types.}  \label{tab:EPFO FuzzQE}
\begin{tabular}{lllllllllll}
\toprule
Algorithm & 1p & 2p & 3p & 2i & 3i & ip & pi & 2u & up & AVG. \\ \midrule
FuzzQE & 43.19 & 9.29 & 7.20 & 24.35 & 29.42 & 12.26 & 16.76 & 12.30 & 7.57 & 18.04 \\ \midrule
MAML & 41.07 & 8.65 & 6.95 & 21.91 & 26.96 & 11.08 & 15.79 & 10.92 & 6.90 & 16.69 \\ \midrule
MAMO(O) & \textbf{43.62} & \textbf{9.42} & \textbf{7.31} & \textbf{26.92} & \textbf{33.03} & \textbf{13.48} & \textbf{17.96} & \textbf{13.11} & \textbf{8.08} & \textbf{19.21} \\ \bottomrule
\end{tabular}
\end{table}

\begin{table}[t]
\centering
\caption{The MRR result(\%) of ConE in all EPFO query types.} \label{tab:EPFO ConE}
\begin{tabular}{lllllllllll}
\toprule
Algorithm & 1p & 2p & 3p & 2i & 3i & ip & pi & 2u & up & AVG. \\ \midrule
ConE & 40.79 & 4.64 & \textbf{4.37} & 19.53 & 23.69 & 4.79 & 16.08 & 12.56 & \textbf{3.98} & 14.49 \\ \midrule
MAML & 40.08 & 2.08 & 2.23 & 15.73 & 17.37 & 2.69 & 8.70 & 11.64 & 1.68 & 11.36 \\ \midrule
MAMO(O) & \textbf{41.06} & \textbf{4.83} & 4.20 & \textbf{27.65} & \textbf{36.64} & \textbf{5.12} & \textbf{18.27} & \textbf{13.43} & 3.90 & \textbf{17.23} \\ \bottomrule
\end{tabular}
%\vskip-2em
\end{table}

\subsection{Results on EPFO and EFO-1 Queries}

The outcome of the extended experiments that are conducted on the EPFO query types and EFO-1 query types is shown in Table~\ref{tab:EPFO formulas LogicE} and Table~\ref{tab:all formulas LogicE}. The Binary Input categorization is omitted in the EPFO since it is the same as the Input.

In both settings, the directly applied MAML performs much worse than the original LogicE in complex structured queries such as pi and pni even though its performance in the 1p/2p/3p queries is comparable to the original model. This result shows that applying MAML at the task level fails to combinatorially generalize.

As for the result of the MAMO algorithm, on average of all query types, we have found it surpasses the original LogicE by around 0.7 percent in the EFO-1 settings and around 1 percent in the EPFO settings by its best variant. Additionally, MAMO outperforms original LogicE in all query types in the EPFO setting and the great majority of query types in the EFO-1 query setting, indicating the tremendous combinatorial generalizability of MAMO.

\subsection{Other Backbone Models}
We investigate the effects of backbone models by applying MAMO on FuzzQE and ConE.
For convenience, we choose MAMO(O) as the variant for evaluation since it is the best in multi-hop and EPFO, and second-best in the EFO-1 when applied to LogicE.
% Among the many variants of MAMO, the Output variant achieves the best in multi-hop and EPFO, and second-best in the EFO-1 when applied to LogicE, thus we choose it to represent the MAMO algorithm when applied to FuzzQE and ConE.
These two experiments are conducted in the EPFO setting for the reason we have discussed above, and the result is shown in Table~\ref{tab:EPFO FuzzQE} and Table~\ref{tab:EPFO ConE}, respectively. These results show that our algorithm is really model agnostic, it improves the performance of FuzzQE by 1.2 percent and ConE by 2.7 percent.

\section{Conclusion}
In this paper, we propose Model Agnostic Meta Operator (MAMO) as an appropriate solution to train complex query answering models when the amount of data is strictly limited. Our research shows that the operator types, rather than query types play a pivotal role in promoting generalizability. Moreover, we investigate several methods to categorize the operator types and show the output is the most effective one that can improve the performance of existing models. More detailed categorizations are worth discussing in future work.

\subsubsection*{Acknowledgments}
The authors of this paper were supported by the NSFC Fund (U20B2053) from the NSFC of China, the RIF (R6020-19 and R6021-20) and the GRF (16211520 and 16205322) from RGC of Hong Kong. We also thank the support from the UGC Research Matching Grants (RMGS20EG01-D, RMGS20CR11, RMGS20CR12, RMGS20EG19, RMGS20EG21, RMGS23CR05, RMGS23EG08). 

\bibliography{ref}
\bibliographystyle{tmlr}

% \appendix
% \section{Appendix}

\end{document}